\pdfoutput=1

\documentclass[11pt]{article}

\usepackage[final]{acl}

\usepackage{times}
\usepackage{latexsym}

\usepackage[T1]{fontenc}

\usepackage[utf8]{inputenc}

\usepackage{times}
\usepackage{amsthm}
\usepackage{fontawesome}
\usepackage{hyperref}

\usepackage{latexsym}
\usepackage{graphicx}
\usepackage{color}
\usepackage{caption}
\usepackage{subcaption}
\usepackage{bbm}
\usepackage{url}
\usepackage{amssymb}
\usepackage{amsmath}
\usepackage{multirow}
\usepackage{booktabs}
\usepackage{algorithm}
\usepackage{algpseudocode}
\usepackage{pifont}
\usepackage{wrapfig}

\newcommand{\algo}{\textsc{FREE}}
\title{\includegraphics[height=1.3em]{latex/figures/3d-render-flash-lightning-sale-thunder-bolt-icon.pdf} \algo{}: Fast and Robust Vision Language Models with Early Exits}

\author{Divya Jyoti Bajpai and Manjesh Kumar Hanawal\\
   Department of IEOR, IIT Bombay \\
  \texttt{\{divyajyoti.bajpai, mhanawal\}@iitb.ac.in}}

\begin{document}
\maketitle
\begin{abstract}
In recent years, Vision-Language Models (VLMs) have shown remarkable performance improvements in Vision-Language tasks. However, their large size poses challenges for real-world applications where inference latency is a concern. To tackle this issue, we propose employing Early Exit (EE) strategies in VLMs. However, training exit classifiers in VLMs is challenging, particularly with limited labeled training data. To address this, we introduce \algo{}, an adversarial training approach within a GAN-based framework. Here, each exit consists of a transformer layer and a classifier. The transformer layer is adversarially trained to produce feature representations similar to the final layer, while a feature classifier serves as the discriminator. Our method focuses on performing input-adaptive inference that increases inference speed with minimal drop in performance. Experimental results demonstrate the effectiveness of our approach in enhancing accuracy and model robustness by mitigating overthinking and the phenomenon of {\it mid-crisis} that we highlight. We experimentally validate that our method speeds up the inference process by more than $1.51\times$ while retaining comparable performance. The source code is available at \noindent\href{https://github.com/Div290/FREE}{\faGithub}.
\end{abstract}
\begin{figure*}[ht]
    \centering
    \includegraphics[scale = 0.42]{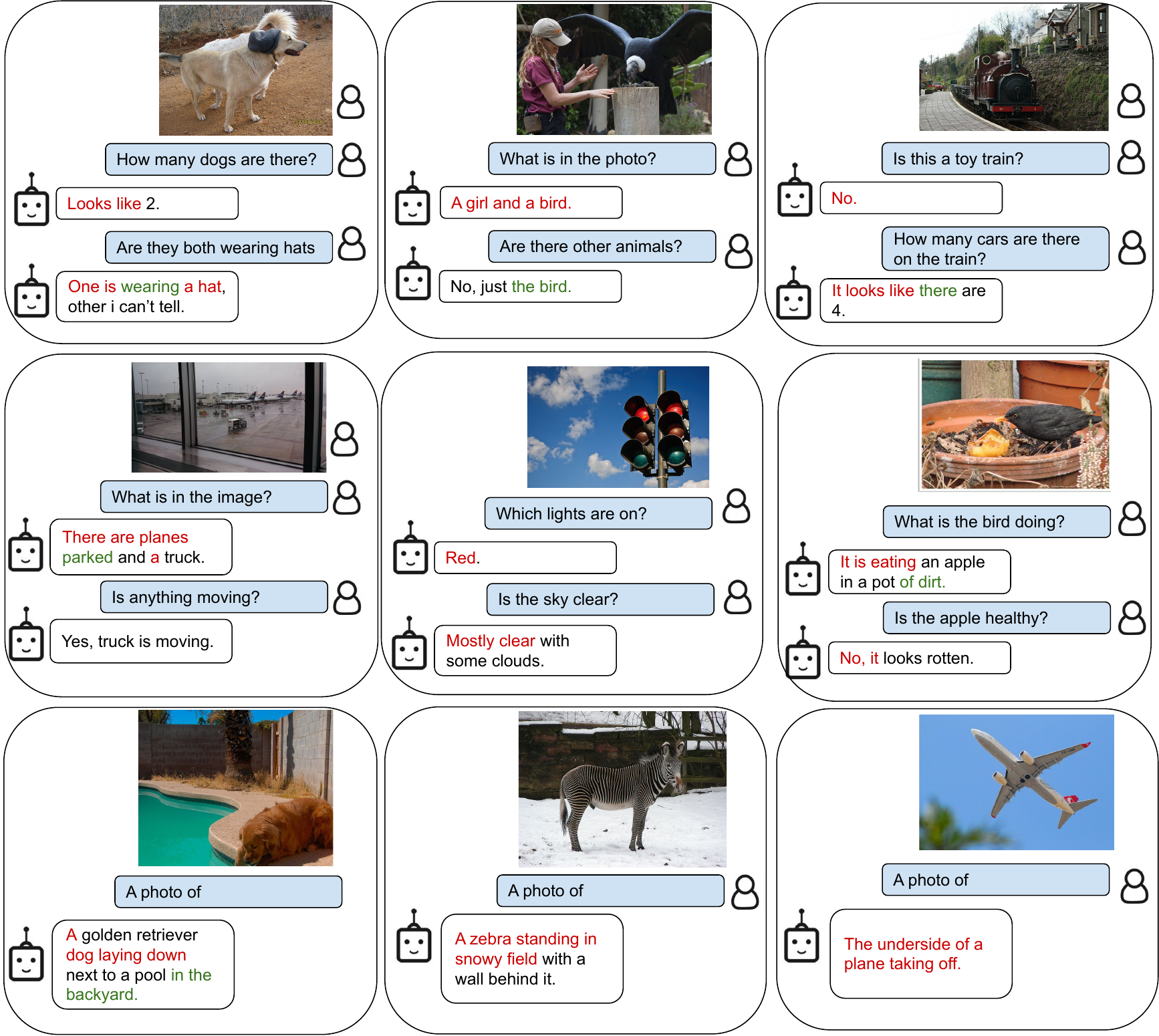}
    \caption{This figure provides some example outputs of \algo{} ViT-g OPT\textsubscript{2.7B}. The different colours show the difficulty levels of tokens in captions. Red: Easy to predict and predicted at initial (1-12) layers. Green: Mediocre hard, predicted at intermediary (13-24) layers. Black: Hard to predict, predicted at deeper (25-32) layers.}
    \label{fig:captions}
\end{figure*}

\section{Introduction}
Vision-language pre-training (VLP) has evolved significantly with the emergence of sophisticated pre-trained Vision Language Models (VLMs). These models have consistently pushed the performance boundaries across various vision-language tasks. However, their demanding computational requirements and inference latency pose challenges for real-time applications. Several models, such as BLIP-2 \cite{li2023blip}, MiniGPT \cite{zhu2023minigpt} etc., leverage off-the-shelf large-scale pre-trained models as building components with their parameters frozen. This reduces VLMs training parameters but makes them extremely slow during inference, as highlighted in \cite{bajpai-hanawal-2024-capeen} leading to higher inference time. 

The use of the Language Model (LM) component with frozen parameters not only makes VLM susceptible to overthinking but also to another phenomenon that we term \textit{mid-crisis} \cite{elhoushi2024layer}. This phenomenon occurs when intermediate layers suffer performance drops due to the search for irrelevant features. While initial layers capture shallow representations and syntactic information, and deep layers learn semantic-fusion relations \cite{fei2022deecap}, intermediary layers tend to capture dataset patterns that degrade their performance, even losing the information learned by initial layers, and the model has to regain the lost information again in deeper layers.

\begin{figure*}
    \centering
    \begin{subfigure}{0.39\textwidth}
        \includegraphics[scale=.35]{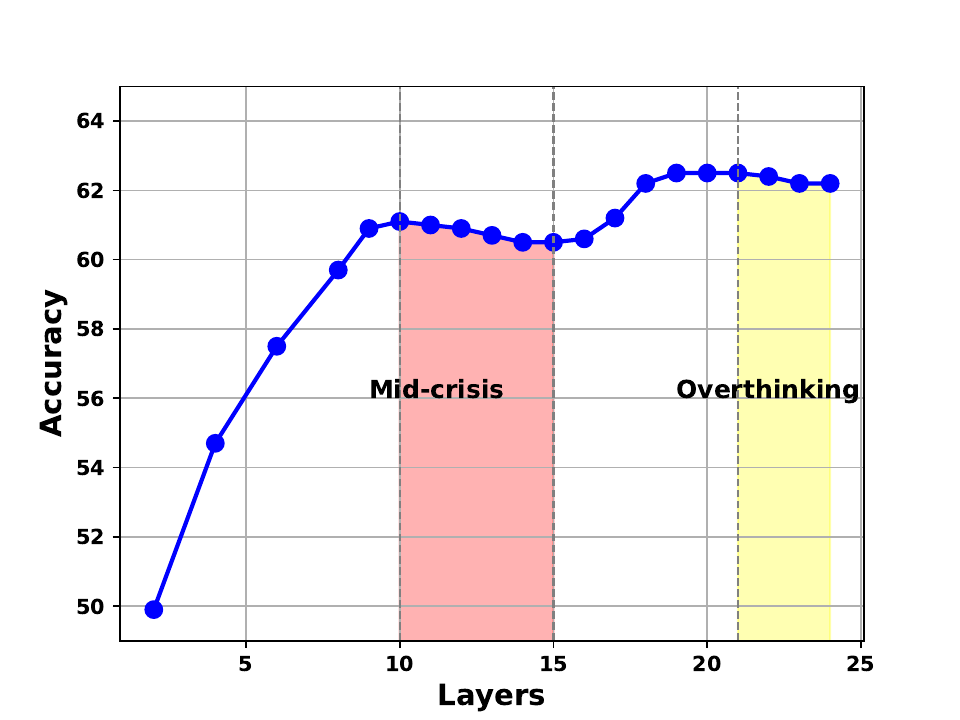}
        \label{fig:speedvsacc}
    \end{subfigure}
    \begin{subfigure}{0.39\textwidth}
        \includegraphics[scale=.35]{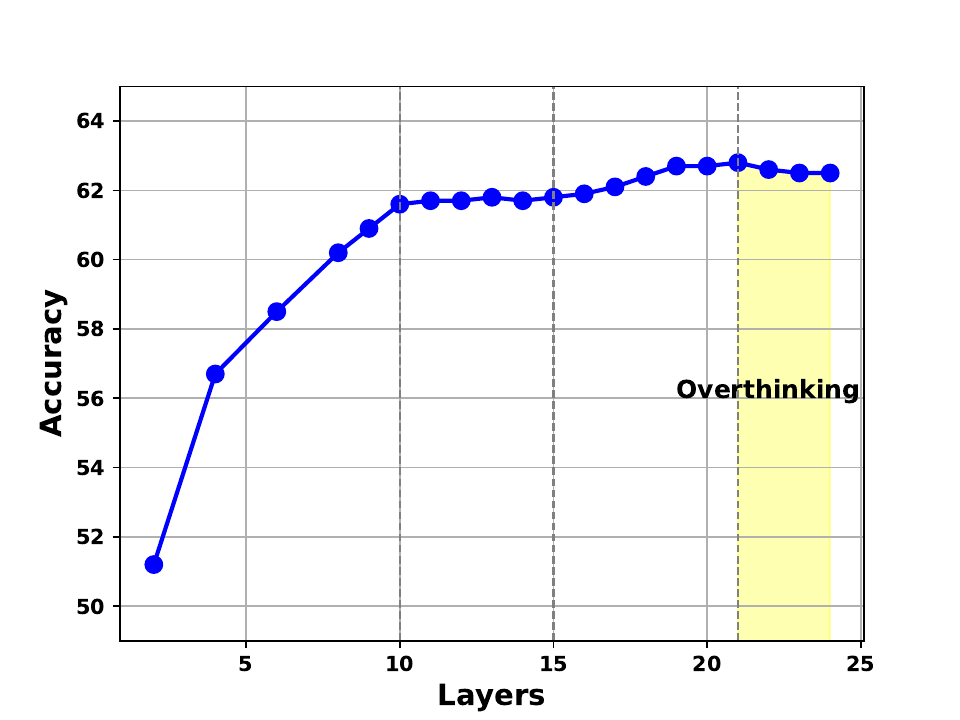}
        \label{fig:confidence_dist}
    \end{subfigure}
    \caption{Left: VQA accuracy on the VQAv2 dataset with BLIP-2-ViT-g-FlanT5\textsubscript{XL} showcasing mid-crisis and overthinking. Right: Our training process improves the mid-crisis. Overthinking can then be solved using EE.}
    \label{fig: mid-level}
\end{figure*}

We illustrate this phenomenon in the left figure of Fig.~\ref{fig: mid-level} and Sec.~\ref{sec: motivation}, showcasing accuracy on the VQAv2 (visual question answering) \cite{das2017visual} dataset across different exits when trained with dedicated classifiers as in vanilla early exit setup for the intermediate layers. This raises the question: how can we mitigate mid-crisis and overthinking to enhance the accuracy and efficiency of VLMs?

We address this using Early Exit (EE) techniques \cite{xin2020deebert, zhou2020bert, zhu2021leebert}, an input-adaptive method that reduces computational costs by bypassing certain layers while preserving the knowledge encoded in large-scale models. Since real-world datasets contain both ``easy'' and ``hard'' samples, EE ensures adaptive computation per sample.

However, applying EE to VLMs presents challenges: (1) Exit classifiers introduce significant overhead---e.g., a single exit for OPT\textsubscript{2.7B} \cite{zhang2022opt} decoder adds approximately 130M parameters; (2) Training these exits requires substantial labeled data, limiting EE adoption in zero-shot VLMs that require minimal fine-tuning.

We propose \algo{} \underline{F}ast and \underline{R}obust Vision-Language Models with \underline{E}arly \underline{E}xits. This efficient EE training framework minimizes training costs while maintaining accuracy. Our method employs a Generative Adversarial Network (GAN)-based \cite{creswell2018generative} framework, leveraging the pre-trained VLM outputs to align feature representations between intermediate exits and the final layer. Unlike cosine similarity, which hampers generalization, the adversarial setup improves feature consistency, ensuring robust exit predictions.

\begin{figure*}
    \centering
    \includegraphics[scale = 0.41]{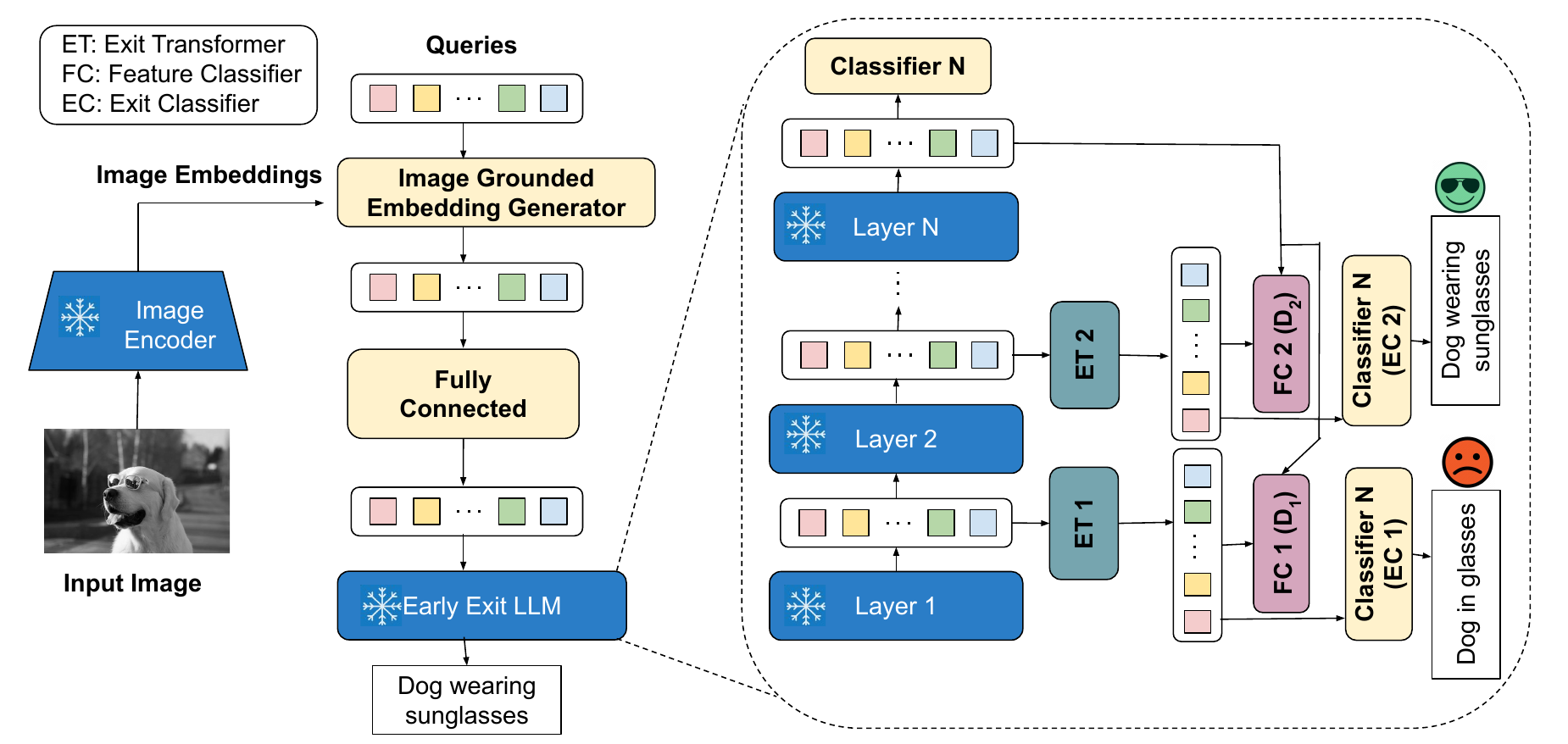}
    \caption{This figure illustrates our model's architecture. The Q-Former output and previous tokens pass through the backbone. The final layer and ET outputs go to feature classifiers, refining ET to match final representations. During inference, the final classifier is used, and confident exits generate the caption.}
    \label{fig:main_figure}
\end{figure*}

Our method attaches exits and {\it Feature Classifiers} (FCs) to intermediary layers of the decoder of the VLM. Each exit consists of a single {\it exit transformer} (ET) and an {\it exit classifier} (EC). The exit transformer is a replica of the layers in the decoder LM of the VLM. They are used as generators and feature classifiers as discriminators in a GAN-based setup, as shown in Fig.~\ref{fig:main_figure}. The task of the feature classifier is to correctly classify if the input is from the exit or final layer, and the task of the exit transformer is to generate representations similar to the final layer to fool the feature classifier of that exit. 

As the exit transformer is trained to generate representations similar to that of the final layer, we can use the final layer classifier at all exits as an EC with frozen parameters, which are used to map the outputs of the exit transformer to class probabilities. As the size of ETs is significantly smaller than that of ECs, it substantially reduces the number of parameters. In this way, a single LM layer attached to the exits helps produce similar feature representations and reduces the training parameters by utilizing a frozen final layer classifier. By attaching EEs, our method reduces the chances of mid-crisis or overthinking and makes the inference process faster. Fig.~\ref{fig:captions} shows how our method can speed up inference while maintaining comparable performance by using the EE methods. 



Adversarial training methods are prone to \textit{catastrophic forgetting} \cite{kirkpatrick2017overcoming, ryu2022knowledge} and \textit{mode collapse}, which can hinder exit training. To address these challenges, we propose both supervised and unsupervised strategies. In the supervised setting, when a small labeled dataset is available, we use \textit{hard labels} to stabilize training, preventing catastrophic forgetting and avoiding local optima. In the unsupervised case, when labeled data is unavailable, we utilize \textit{soft labels} from vanilla VLM or generate high-quality synthetic labels using \textit{CapFilt} \cite{li2022blip}, as employed in BLIP and BLIP-2.

Our method remains effective across different scenarios. When a labeled dataset is available, it is used to mitigate catastrophic forgetting and mode collapse. If no labeled data is present, we employ knowledge distillation to solve this issue. When both labeled data and annotations are unavailable but computational resources are accessible, we generate synthetic labels using CapFilt. By providing solutions for all data constraints, our approach ensures stable and effective exit training while enhancing efficiency and generalization. In summary, our contributions are as follows:

\begin{itemize}
    \item We introduce an EE strategy named \algo{} for VLMs to effectively mitigate inference latency by reducing unnecessary computations inherent in their large-scale architecture. 
    
    \item We propose an efficient training strategy to improve performance at EE classifiers. \algo{} emulates the behaviour of the final layer at the exits through adversarial learning. This reduces the need for labeled training datasets.
    

    \item  Our model reduces the number of trainable parameters of the exits by reutilizing the frozen final layer classifier at the exits.

    \item  We experiment both qualitatively (see Fig.~\ref{fig:captions} (\textbf{recommended})) and quantitatively on various tasks such as image captioning, visual question-answering and visual dialogue dataset. Our method provides inference speed $>1.51\times$ with comparable accuracy than vanilla VLM inference. We show the robustness of our method in Appendix \ref{sec: robustness}.
\end{itemize}

\section{Related works}
We discuss the VLPs with LM components and EE strategies related to our work below. 

{\bf Vision-language Pre-training:}
Different model architectures have been proposed for specific downstream tasks in VLPs, including dual-encoder architectures \cite{radford2021learning, jia2021scaling}, encoder-decoder architectures \cite{cho2021unifying, wang2021simvlm, chen2022pali}. Various pre-training objectives have also been introduced, such as image-text contrastive learning \cite{radford2021learning,yao2021filip,li2021align, li2022blip}, the image-text matching \cite{ju2021learning, li2022blip, bao2021vlmo}, and masked language modeling \cite{li2021align, li2022blip, yu2022coca, wang2022image}. However, these end-to-end models are inflexible to leverage readily available pre-trained models, such as LLMs \cite{brown2020language, zhang2022opt,chung2024scaling}. 

Recent approaches in vision-language pre-training have adopted the strategy of utilizing off-the-shelf pre-trained models and keeping them frozen during training. Some methods freeze only the image encoder \cite{chen2020uniter, li2020oscar, zhang2021vinvl}, and recent LiT \cite{zhai2022lit} which uses a frozen pre-trained encoder for CLIP \cite{radford2021learning} pre-training, while others freeze the language model to leverage knowledge from language-only pre-trained models for vision-to-language generation tasks \cite{tsimpoukelli2021multimodal, alayrac2022flamingo, chen2022visualgpt, manas2022mapl, tiong2022plug, guo2022images}. The primary challenge lies in aligning visual features with text space. To address this challenge, techniques like Frozen \cite{tsimpoukelli2021multimodal} finetune image encoders or insert new cross-attention layers into language models to incorporate visual features. BLIP-2 \cite{li2023blip} employs both frozen image encoders and language models for vision-language tasks, achieving strong performance.

{\bf Early Exits:} To minimize inference latency in deep neural networks, BranchyNet \cite{teerapittayanon2016branchynet} introduced attaching exits classifiers at intermediary layers. This concept was extended by Shallow-deep \cite{kaya2019shallow}, which effectively determines when to exit based on confidence distribution at each exit classifier. Approaches like \cite{huang2017multi, yang2020resolution, han2023dynamic} improve EEs for image tasks by dynamically choosing different depths for different regions of the image. Approaches like \cite{phuong2019distillation} have employed knowledge distillation for image classification.

For NLP tasks, several early exit frameworks have emerged \cite{xin2020deebert, liu2021towards, zhou2020bert, liu2020fastbert, wang2019dynexit, wang2020accelerating, zhu2021leebert, ji2023early, zhang2022pcee, 10.1145/3639856.3639873, 10622954, bajpai2024distributed, bajpai-hanawal-2024-ceebert, bajpai-hanawal-2024-capeen, bajpai-hanawal-2024-dadee, bajpaibeem},  primarily built on the BERT backbone. DeeCap \cite{fei2022deecap} introduces EE for image captioning tasks, employing an imitation network to replicate outputs from transformer layers. MuE \cite{tang2023you} applies EE to OFA \cite{wang2022ofa}, a VLM tailored for multi-modal applications. CapEEN \cite{bajpai-hanawal-2024-capeen} makes the EEs robust to noise by adapting to the exiting threshold.

The key differences in our work are: 1) We employ adversarial training for efficient learning of EE models. 2) Our method can work under both supervised and unsupervised setups by utilizing the zero-shot capabilities of the VLMs, while previous methods require a good amount of high-quality labeled training data.

\section{Methodology}
We begin with a pre-trained Vision-Language Model (VLM) comprising three key components: a visual encoder, a projection layer, and a language model (LM). The visual encoder processes the input image, which may be a standalone image encoder \cite{wang2022ofa} or include an image-grounded text encoder (e.g., Q-Former in BLIP-2). The projection layer then transforms the visual features to align with the LM’s input space which is passed as an input to the decoder. As autoregressive decoding requires multiple forward passes of LM, our focus will primarily be on optimizing the LM component for efficient inference.

We assume that LM consists of $N$ layers, and we attach exits to the $K$ chosen layers. Each exit consists of one Exit Transformer (ET) layer with the same configuration as one LM layer and an Exit Classifier (EC). The exit layer is such the parameters of the ET are trainable, and those of EC are frozen. Before indulging into architectural details, we motivate our method by highlighting the major concerns on the well-known BLIP-2 model. 
\subsection{Motivation}\label{sec: motivation}
In Fig.~\ref{fig: mid-level}, we show the VQA accuracy for various layers on the VQAv2 dataset on the BLIP-2 model. In the left side of the figure, we show the performance of vanilla EE methods where just an EC is attached to the intermediate layer output. As seen, performance dips at the middle layers after the initial improvement. This is due to the LM component, which is of large size and kept frozen during training. During pre-training, BLIP-2 aligns the features of text and images using the Q-former, the querying transformer. However, the Q-former provides the image-grounded text embeddings to the LM component in such a way that it produces high-quality predictions only at the final layer and not the intermediary layer. 

To address this, we enhance exits by incorporating a transformer layer instead of relying solely on an EC. This layer mimics the final layer’s behavior, enabling intermediate exits to access deep-level representations. By doing this, we are not directly passing the information of intermediate layers to the classifiers instead we first enhance the information through ETs and then pass the hidden representations to classifiers which improves the performance of exits. The right side of Fig.~\ref{fig: mid-level} demonstrates how our approach reduces mid-crisis by leveraging deeper layer information. Overthinking is further alleviated by performing inference through EEs using thresholds.

Previous EE methods that attach a trainable classifier at exits introduce substantial parameter overhead, given the large vocabulary size. For instance, a single classifier added to BLIP-2 with OPT\textsubscript{2.7B} contributes $130M$ trainable parameters, and $7$ exits would scale this up to $900M$ parameters. In contrast, our method only trains the LM layer at each exit, adding $63M$ parameters per exit and $588M$ for $7$ exits—reducing the trainable parameters by approximately $52$\% (see Appendix \ref{sec:parameter_count}). This efficiency is achieved by using the final layer classifier and training only the LM layer. These issues also translate to other VLMs such as MiniGPT \cite{zhu2023minigpt} and InstructBLIP \cite{dai2024instructblip} due to similar architectures.

\noindent
We next discuss our method which consists of two parts: 1) backbone finetuning and 2) exit training.

\subsection{Backbone finetuning}
The backbone is fine-tuned using the cross-entropy loss between the predicted token and the ground truth token. The loss function for fine-tuning could be formulated as:
\begin{equation}
    \mathcal{L}(I) = \sum_{t = 1}^T\text{log}P_N(y_t^*|y_{1:t-1}^{*}, I)
\end{equation}
where $I$ denotes the input image, $y_{1:T}^*$ is the true caption and $T$ is the caption length. $P_N$ denote the probability output from the final layer. In this step, the backbone learns to produce high-quality features at the final layer and a classifier $C_N$ to map the feature representations at the final layer to class probabilities. Note that $C_N$ is part of the backbone. Once the fine-tuning is complete, we freeze the parameters of the backbone. This is done to maintain the optimality of the backbone.

\subsection{Exits training}
After fine-tuning the backbone, we attach $K$ exits to the LM component of the VLM.  We denote the set of exit indices by $[K]$. At each exit, we use a feature classifier $D_i$ that discriminates the feature representations of the transformer layer of the $i$th exit from that of the final layer. Specifically, it provides a score to an input feature representation, whether it is from the final layer.
We have a separate feature classifier for every exit as feature representations at different exits can differ. 

In our setup, the feature classifier acts as a discriminator, and the exit transformer layer as a generator; the goal of the transformer layer is to generate feature representations similar to that of the final layer. We train them alternately as in the original GAN framework. This training problem can be set up as an unconstrained optimization problem. Let $E_i$ denote the transformer layer in the $i$th exit. The feature classifier loss for an exit $i \in [K]$ and an input image $I$ could be formulated as:
\begin{align}
    \mathcal{L}_i^{fc}(h_t^i, h_t^N \mid y_{1:t-1}^*, I) = 
    -\text{log}D_i(h_t^N \mid y_{1:t-1}^*, I) \nonumber \\ 
    - \text{log}(1 - D_i(E_i(h_t^i \mid y_{1:t}^*, I)))\nonumber
\end{align}
where $h_t^i$ is the feature (hidden) representation at $i$th layer and $h_t^N$ is the feature representation at the final layer of the LM for the $t$th token in the sentence. The overall loss across all exits could be written as $\mathcal{L}^{fc} = \sum_{i\in [K]}\mathcal{L}_i^{fc}$. It simultaneously updates the feature classifiers across all the exits.

The generator loss for the transformer layer in $i$th exit could be formulated as: 
\begin{equation}
    \mathcal{L}_i^{gen}(h_t^i|y_{1:t-1}^*,I) = -\text{log}D_i(E_i(h_t^i|y_{1:t-1}^*, I)))\nonumber
\end{equation}
However, because the weights of the transformer layer of the exits are untied from the original backbone, they can face the issue of catastrophic forgetting or mode collapse. To circumvent these issues, we utilize the small labeled dataset to fine-tune the backbone. It guides the model to correct the learning trajectory and not let it get stuck to the local minima. The labeled data not only removes the issue of catastrophic forgetting but also helps in reducing overthinking as exits mimic the final layer and learn from the hard labels. The loss could be written as:
\begin{equation}
    \mathcal{L}_i^{CE}(I, y_{1:t-1}^{*}) =\text{log}P_i(y_t^*|y_{1:t-1}^{*}, I)
\end{equation}
where $P_i$ denotes the probability score at the $i$th exit. The loss at exit $i$ will be $\mathcal{L}_i = \mathcal{L}_i^{CE}+\mathcal{L}_i^{gen}$. The overall loss for exits training will be $\mathcal{L} = \sum_{i \in [K]} \mathcal{L}_i$. 
After this step, the backbone is learned with attached exits.

\subsection{Unsupervised setup}
Recall from the previous section that labeled data was utilized to reduce the issue of catastrophic forgetting and mode collapse. As the VLMs has good zero-shot performance, we can utilize it to either distill the knowledge at the final layer or create a small set of pseudo labels to fulfill the requirements of the labeled dataset. We provide two methods for unsupervised learning. 

{\bf Using Knowledge Distillation:} In this case, we can directly utilize the soft labels from the final layer to distill the knowledge to the exits. The knowledge distillation loss for the $i$th layer could be formulated as:
\begin{equation}
    \mathcal{L}_i^{KL} = KL(p_t^i, p_t^N)
\end{equation}
where $p_t^i = C_N(E_i(h_t^i|y_{1:t-1}^{*}, I))$, $p_t^N = C_N(h_t^N|y_{1:t-1}^{*}, I)$ and $KL$ is the KL-divergence loss defined as $KL(p_t^i, p_t^N) = \sum_{v\in\mathcal{V}}p_t^i\log\frac{p_t^i}{p_t^N}$ where $\mathcal{V}$ is the vocabulary. We can train the exits by replacing the $\mathcal{L}_i^{CE}$ by $\mathcal{L}_i^{KL}$. Then the overall loss for exit $i$ is $\mathcal{L}_i = \mathcal{L}_i^{KL}+\mathcal{L}_i^{gen}$. 

This method also utilizes the zero-shot capabilities of the VLMs model. However, this method has a slightly lower performance than the CapFilt method proposed next, still, it comes with lower computational cost and has comparable performance to vanilla VLM inference. 

{\bf Using CapFilt:} CapFilt \cite{li2022blip, li2023blip} is a method that is used in both the original BLIP and BLIP-2 models to generate high-quality synthetic captions. We use similar ideas to generate the labeled dataset. In this step, a sample is passed through the BLIP-2 model, which then provides us with $10$ possible captions for the given samples. We then use the CLIP ViT-L/14 model to rank the synthetic captions based on the image-text similarity produced by the CLIP model. We then keep the top-2 captions and keep them as synthetic captions that can be later utilized for training the exits by treating the synthetic captions as true captions. Creating synthetic captions using the CapFilt is more accurate but computationally heavy \cite{li2022blip}. Hence it is recommended when the computational resources are available.  

\subsection{Inference}
We perform captioning in an autoregressive manner. This entails making a token-by-token prediction for a given image, where the layer at which the token is predicted is determined by the confidence score $S_i = \max_{v\in\mathcal{V}}C_N(E_i(h_t^i|y_{1:t-1}, I))(v)$ where $\mathcal{V}$ is the vocabulary. The input to the decoder is processed sequentially through the decoder layers until the confidence score $S_i$ is greater than a predefined threshold value $\alpha$. The inference starts with the begin of the sentence token and the next token is predicted either at the exits or at the final layer. The inference process stops when the end of the sentence token is predicted either at intermediary layers or at the final layer. Note that if the prediction confidence is below $\alpha$ for all the exits, then the sample is predicted at the final layer, irrespective of the confidence in the prediction.


\section{Experiments}
\begin{table*}[]
\small
\centering
\begin{tabular}{lcccccccccc}
\hline
\multicolumn{1}{l|}{}                    &                                                               & \multicolumn{8}{c}{NoCaps Zero-shot}                                                                                  &               \\
\multicolumn{1}{l|}{Models}              & \begin{tabular}[c]{@{}c@{}}\#Train\\ Params\end{tabular} & \multicolumn{2}{c}{in-domain}  & \multicolumn{2}{c}{near-domain} & \multicolumn{2}{c}{out-domain} & \multicolumn{2}{c}{full-dataset} & Spd         \\
\multicolumn{1}{l|}{}                    &                                                               & C              & S             & C               & S             & C              & S             & C               & S              &               \\ \hline
\multicolumn{1}{l|}{Vin VL}              & 345M                                                          & 102.9          & 14            & 94.8            & 13.7          & 88.1           & 11.9          & 95.1            & 13.2           & -             \\
\multicolumn{1}{l|}{BLIP}                & 446M                                                          & 114.9          & 15.2          & 110.6           & 14.6          & 114.8          & 14.3          & 112.8           & 14.7           & -             \\
\multicolumn{1}{l|}{SimVLM}              & 1.4B                                                          & 113.7          & 14.9          & 110.6           & 14.2          & 114.6          & 14.4          & 112.1           & 14.3           & -             \\
\multicolumn{1}{l|}{BLIP-2 ViT O\textsubscript{2.7B}}    & 1.1B                                                          & 123.0            & 15.8          & 117.8           & 15.4          & 123.2          & 15.0            & 119.6           & 15.4           & 1.07$\times$             \\
\multicolumn{1}{l|}{BLIP-2 ViT FT5\textsubscript{XL}} & 1.1B                                                          & 123.7          & 16.3          & \textbf{120.2}  & \textbf{15.9} & 124.8          & 15.1          & 121.6           & 15.8           & 1.00$\times$             \\ \hline
\multicolumn{11}{c}{\textit{Early Exit models}}                                                                                                                                                                                                                          \\ \hline
\multicolumn{1}{l|}{DeeBLIP}             & 1.8B                                                          & 115.2          & 15.3          & 111.5           & 14.7          & 115.4          & 14.5          & 112.4           & 14.5           & 1.41$\times$          \\
\multicolumn{1}{l|}{PABEE-BLIP}          & 1.8B                                                          & 117.7          & 15.4          & 114.2           & 14.8          & 117.6          & 14.7          & 112.9           & 14.6           & 1.29$\times$          \\
\multicolumn{1}{l|}{LeeBLIP}             & 1.8B                                                          & 119.4          & 15.5          & 115.8           & 14.8          & 120.1          & 14.9          & 116.3           & 15.1           & 1.38$\times$          \\
\multicolumn{1}{l|}{MuE}                 & 1.8B                                                          & 118.1          & 15.4          & 115.3           & 14.8          & 118.7          & 14.8          & 114.8           & 14.9           & 1.44$\times$          \\ \hline
\multicolumn{1}{l|}{\algo{} ViT O\textsubscript{2.7B}}       & 1.5B                                                          & 122.7          & 15.7          & 118.1           & 15.5          & 123.9          & 15.1          & 119.9           & 15.6           & \textbf{1.63}$\times$ \\
\multicolumn{1}{l|}{\algo{} ViT FT5\textsubscript{XL}}    & 1.4B                                                          & \textbf{124.3} & \textbf{16.5} & 120.0             & \textbf{15.9} & \textbf{125.5} & \textbf{15.4} & \textbf{122.7}  & \textbf{16.1}  & 1.51$\times$          \\ \hline
\end{tabular}
\caption{Results on the Nocaps dataset during zero-shot transfer when the model is trained on the COCO dataset across various domains. Spd is the speedup achieved by the model. O\textsubscript{2.7B} is OPT\textsubscript{2.7B} and FT5\textsubscript{XL} is FlanT5\textsubscript{XL}.}
\label{tab: res_unsup_cap}
\end{table*}

\begin{table}
\small
\centering
\begin{tabular}{lcccc}
\hline
\multirow{2}{*}{Models} & \multirow{2}{*}{\begin{tabular}[c]{@{}c@{}}\#Train\\ Params\end{tabular}} & \multirow{2}{*}{\begin{tabular}[c]{@{}c@{}}VQAv2\\ train\end{tabular}} & \multirow{2}{*}{\begin{tabular}[c]{@{}c@{}}VQAv2\\ test\end{tabular}} & \multirow{2}{*}{Spd} \\
                        &                                                                               &                                                                        &                                                                       &                          \\ \hline

\multicolumn{5}{c}{\textit{Without Exits}}                                                                                                                                                                                                                                      \\ \hline                        
ALBEF                   & \multicolumn{1}{c|}{314M}                                                     & 72.3                                                                   & 71.5                                                                  & -                     \\
BLIP                    & \multicolumn{1}{c|}{385M}                                                     & 73.9                                                                   & 72.1                                                                  & -
\\
OFA                     & \multicolumn{1}{c|}{930M}                                                     & 75.7                                                                   & 75.6                                                                  & -
\\
Flamingo80B             & \multicolumn{1}{c|}{10.6B}                                                    & 77.9                                                                   & 78.1                                                                  & -
\\
BLIP-2 V-O        & \multicolumn{1}{c|}{1.2B}                                                     & 78.3                                                                   & 78.5                                                                  & 1.07$\times$                     \\
BLIP-2 V-F     & \multicolumn{1}{c|}{1.2B}                                                     & 78.8                                                          & 78.7                                                                  & 1.00$\times$                     \\ \hline
\multicolumn{5}{c}{\textit{Early Exit models (on BLIP-2)}}                                                                                                                                                                                                                                      \\ \hline
DeeBLIP              & \multicolumn{1}{c|}{1.9B}                                                     & 75.4                                                                   & 75.9                                                                  & 1.52$\times$                     \\
PABEE-BLIP              & \multicolumn{1}{c|}{1.9B}                                                     & 77.4                                                                   & 77.1                                                                  & 1.39$\times$                     \\
LeeBLIP             & \multicolumn{1}{c|}{1.9B}                                                     & 78.1                                                                   & 77.8                                                                  & 1.65$\times$                     \\  \hline
\algo{}-V-O           & \multicolumn{1}{c|}{1.6B}                                                     & 78.7                                                                   & 79.0                                                            & \textbf{1.77$\times$}            \\
\algo{}-V-F        & \multicolumn{1}{c|}{1.5B}                                                     & \textbf{78.9}                                                          & \textbf{79.1}                                                         & 1.71$\times$               \\ \hline
\end{tabular}
\caption{Results of semi-supervised application of our model to Visual-Question Answering tasks.}
\label{tab: res_sup_vqa}
\vspace{-0.5cm}
\end{table}

\textbf{Dataset and Metric:} We evaluate the performance of our method using the COCO \cite{lin2014microsoft} and NoCaps dataset \cite{agrawal2019nocaps} for image captioning. For Visual Question-answering tasks, we utilize the VQAv2 \cite{goyal2017making}, OK-VQA \cite{marino2019ok} and GQA \cite{hudson2019gqa} datasets. For visual dialogue, we use the VisDial \cite{das2017visual} dataset. We report key metrics including BLEU-4 \cite{papineni2002bleu}, METEOR \cite{banerjee2005meteor}, CIDEr \cite{vedantam2015cider} and SPICE \cite{anderson2016spice} scores for captioning. For VQA tasks, we report the VQA accuracy and for the Visual Dialog, we use the Mean
Reciprocal Rank (MRR) \cite{dai2024instructblip}. To effectively consider the speedup, we define the speedup as inverse of the fraction of parameters used for inference on average. The speedup is formulated as:
$$\text{Speedup} =\frac{\text{Total parameters}}{\text{Average number of parameters used}}$$
where the average number of parameters could be written as $\frac{1}{M}\sum_{I=1}^{M}\sum_{i = 1}^{N_I}w_i\times (i+k)\times p$ where $p$ denotes the number of parameters in one layer of the LM component, $N_I$ denotes the number of predicted words in the caption for the image $I$, $M$ denotes the total number of input images and $k=1$ is the number of LM layers in the exit. Total parameters denote the total number of parameters in the backbone. 
The baseline for comparing speedup for BLIP-2 models is BLIP-2 ViT-g FlanT5\textsubscript{XL}. We only compare the speedup of early exiting methods and the BLIP-2 variants. In the Appendix \ref{sec:MiniGPT} and Table \ref{tab:res_minigpt} and \ref{tab:res_minigpt_vqa}, we show the results of the MiniGPT \cite{zhu2023minigpt} model and in Appendix \ref{sec:InstructBLIP} and Table \ref{tab:res_instruct}, we show the results on the InstructBLIP \cite{liu2024visual} model.

\textbf{Training:} In our setup for BLIP-2, we utilize two variations of the BLIP-2 model with the same image encoder (ViT-g/14 \cite{dosovitskiy2020image}) and frozen LLMs that are OPT-2.7B \cite{zhang2022opt} and FlanT5-XL \cite{chung2024scaling}. We use the LAVIS \cite{li2022lavis} library for implementation, training and evaluation. For training, we use the validation split of the datasets. We use $80\%$ of validation split for training and the remaining $20\%$ for development.

We use labels of the validation dataset when the task is semi-supervised, else we just use the samples without labels. First, the backbone is fine-tuned for $10$ epochs with a starting learning rate of 1e-5, which decays by $0.5$ every $3$ epochs. The backbone weights are then frozen post-fine-tuning and exits are attached to the backbone. We train exit weights for a further $3$ epochs. We employ the Adam \cite{kingma2014adam} optimizer and a batch size of $16$. 
Similar to \citet{bajpai-hanawal-2024-dadee}, we use a feature classifier, with one hidden linear layer with a hidden state of size $3072$ and a LeakyReLU activation function.

For the unsupervised tasks, we train the model for $5$ epochs on the validation dataset (without labels) with knowledge distillation from the final layer. Optimizers and learning rates are kept the same as given above. Note that in CapFilt we apply the CapFilt method on the validation dataset (without labels) and generate synthetic labels. After this, we perform a similar procedure by treating the synthetic labels as the true labels as done for the semi-supervised tasks. We do not finetune the backbone in an unsupervised setup. 

\textbf{Inference:} Inference is conducted with a batch size of $1$. We provide the results on the test dataset. For image captioning, we use a prompt as `a photo of'. The threshold is chosen as the best-performing threshold from the set $\{0.5, 0.6, 0.7, 0.8, 0.9, 1.0\}$ on the held-out split of the validation dataset. More details on the hyperparameter setting can be found in Table \ref{tab: hyperparms detail} and in the Appendix \ref{sec: alpha} with the values of hyperparameters.
All experiments were performed with a combination consisting of two NVIDIA RTX A6000 and four NVIDIA GeForce RTX 3080 Ti GPUs.

\textbf{Baselines:}
We establish baseline models for performance evaluation, including vanilla BLIP-2 inference. Additionally, we compare with multimodal models VinVL \cite{zhang2021vinvl}, ALBEF \cite{li2021align}, SimVLM \cite{wang2021simvlm}, OFA \cite{wang2022ofa}, and Flamingo \cite{alayrac2022flamingo}. We also assess state-of-the-art early exit methods originally proposed for the BERT backbone, such as DeeBERT \cite{xin2020deebert}, PABEE \cite{zhou2020bert}, and LeeBERT \cite{zhu2021leebert}, adapted to the BLIP-2 backbone as DeeBLIP, PABEE-BLIP, and LeeBLIP, respectively. DeeBLIP uses confidence-based exiting, PABEE-BLIP employs patience-based exiting, and LeeBLIP combines knowledge distillation from the final layer with hard label learning. Furthermore, we apply the MuE \cite{tang2023you} early exiting method to the BLIP-2 backbone, using exits from the better-performing BLIP-2 variant for our baselines.

\begin{table*}[]
\centering
\small
\begin{tabular}{lcccccccc}
\hline
\multicolumn{1}{c}{\multirow{2}{*}{Model}}   & \multirow{2}{*}{\begin{tabular}[c]{@{}c@{}}\#Total\\ params\end{tabular}} & \multirow{2}{*}{\begin{tabular}[c]{@{}c@{}}VisDial\\test\end{tabular}} & \multicolumn{2}{c}{\multirow{2}{*}{\begin{tabular}[c]{@{}c@{}}VQAv2\\ train   
 test\end{tabular}}} & \multirow{2}{*}{\begin{tabular}[c]{@{}c@{}}OK-VQA\\ test\end{tabular}} & \multirow{2}{*}{\begin{tabular}[c]{@{}c@{}}GQA\\ test\end{tabular}} & \multirow{2}{*}{\begin{tabular}[c]{@{}c@{}}VizWiz\\ test\end{tabular}} & \multirow{2}{*}{Speed} \\
\multicolumn{1}{c}{}                         &                                                                               &                                                                         & \multicolumn{2}{c}{}                                                                              &                                                                        &                                                                     &                                                                        &                        \\ \hline
\multicolumn{9}{c}{\textit{Without exits}}                                                                                                                                                                                                                                                                                                                                                                                                                                                                                                                      \\ \hline
\multicolumn{1}{l|}{Flamingo3B}              & 3.2B                                                                          & -                                                                    & 53.2                                            & 49.4                                            & 41.5                                                                   &                                                          -           & 28.9                                                                   & 1.28$\times$                      \\
\multicolumn{1}{l|}{Flamingo9B}              & 9.3B                                                                          & -                                                                    & 55.7                                            & 51.8                                            & 44.7                                                                   &                                                       -              & 28.8                                                                   & 0.44$\times$                      \\
\multicolumn{1}{l|}{Flamingo80B}             & 80B                                                                         & -                                                                     & 59.1                                            & 56.2                                            & \textbf{50.4}                                                          &                                                    -                 & \textbf{31.5}                                                          & 0.05$\times$                      \\
\multicolumn{1}{l|}{BLIP-2 ViT-g OPT\textsubscript{2.7B}}        & 3.8B                                                                          & 34.1                                                                    & 54.6                                            & 52.0                                              & 31.2                                                                   & 34.2                                                                & 27.0                                                                     &          1.07$\times$             \\
\multicolumn{1}{l|}{BLIP-2 ViT-g OPT\textsubscript{6.7B}}        & 7.8B                                                                          & 37.5                                                                   & 55.9                                            & 53.7                                            & 36.1                                                                   & 36.4                                                                & 27.2                                                                   & 0.52$\times$                      \\
\multicolumn{1}{l|}{BLIP-2 ViT-g FlanT5\textsubscript{XL}}     & 4.1B                                                                          & \textbf{45.9}                                                                    & \textbf{64.9}                                   & \textbf{62.5}                                   & 40.6                                                                   & \textbf{44.5}                                                       & 29.8                                                                   & 1.00$\times$                      \\ \hline
\multicolumn{9}{c}{\textit{Early Exit models   (on BLIP-2 ViT-g FlanT5\textsubscript{XL}})}                                                                                                                                                                                                                                                                                                                                                                                                                                                                                                         \\ \hline
\multicolumn{1}{l|}{DeeBLIP}                 & 4.7B                                                                          & 33.4                                                                    & 41.3                                            & 42.8                                            & 23.4                                                                   & 27.8                                                                & 20.1                                                                   & 1.39$\times$                   \\
\multicolumn{1}{l|}{PABEE-BLIP}            & 4.7B                                                                          & 35.7                                                                   & 49.6                                            & 51.3                                            & 31.2                                                                   & 34.3                                                                & 23.6                                                                   & 1.31$\times$                   \\
\multicolumn{1}{l|}{LeeBLIP}                 & 4.7B                                                                          & 39.1                                                                    & 57.7                                            & 57.1                                            & 37.1                                                                   & 39.7                                                                & 26.4                                                                   & 1.29$\times$                   \\
\multicolumn{1}{l|}{MuE}                     & 4.7B                                                                          & 36.6                                                                    & 55.4                                            & 53.6                                            & 33.7                                                                   & 37.1                                                                & 24.7                                                                   & 1.36$\times$                   \\  \hline
\multicolumn{1}{l|}{\algo{} ViT-g OPT\textsubscript{2.7B}}    & 4.3B                                                                          & 32.3                                                                    & 55.5                                            & 53.4                                            & 35.6                                                                   & 44.7                                                                & 26.8                                                                   & \textbf{1.51}$\times$          \\
\multicolumn{1}{l|}{\algo{} ViT-g FlanT5\textsubscript{XL}} & 4.5B                                                                          & \textbf{45.5}                                                                    & \textbf{64.5}                                   & \textbf{62.1}                                   & \textbf{40.3}                                                          & \textbf{44.0}                                                         & \textbf{29.5}                                                          & {1.45}$\times$                   \\ \hline
\end{tabular}
\caption{Results of the unsupervised Visual-Question Answering and VisDial dataset. For VQA tasks, we report the VQA accuracy and for the visual dialogue, we report the Mean Reciprocal Rank(MRR).}
\label{tab: res_unsup_vqa}
\end{table*}

\begin{table}
\centering
\small
\begin{tabular}{lccccc}
\hline
\multicolumn{1}{l|}{\multirow{2}{*}{Models}} & \multicolumn{5}{c}{COCO Karpathy test}                                 \\
\multicolumn{1}{l|}{}                        & B@4           & C              & S             & M             &    Spd            \\ \hline
\multicolumn{1}{l|}{OFA}                     & 37.5 & 130.3          & 25.2          & 31.1          &              -
\\
\multicolumn{1}{l|}{Flamingo}                & 38.5         & 134.1          & 24.1          & 27.8          &             -
\\
\multicolumn{1}{l|}{SimVLM}                  & 38.6          & 138.3          & 24.8          & 29.8          &              -
\\
\multicolumn{1}{l|}{BLIP-2-V-O}        & 41.7          & 139.8          & 25.5          & 30.5          & 1.07$\times$             \\
\multicolumn{1}{l|}{BLIP-2-V-F}     & 40.4          & 141.5          & 25.2          & 29.1          & 1.00$\times$            \\ \hline
\multicolumn{6}{c}{\textit{Early Exit models}}                                                                                         \\ \hline
\multicolumn{1}{l|}{DeeBLIP}                 & 32.8          & 115.1          & 20.9          & 25.3          & 1.65$\times$          \\
\multicolumn{1}{l|}{PABEE-BLIP}              & 34.2          & 119.8          & 21.4          & 26.2          & 1.45$\times$          \\
\multicolumn{1}{l|}{LeeBLIP}                 & 37.4          & 132.0            & 22.8          & 27.6          & 1.59$\times$          \\
\multicolumn{1}{l|}{MuE}                     & 37.9          & 137.5          & 23.6          & 28.5          & 1.41$\times$          \\
\multicolumn{1}{l|}{\algo{}-V-O}    & \textbf{41.9} & \textbf{142.5} & \textbf{25.2} & \textbf{30.8} & \textbf{1.75}$\times$ \\ \hline
\end{tabular}
\caption{Results of semi-supervised training on the Karpathy test split of the COCO dataset.}
\label{tab: res_sup_coco}
\vspace{-0.5cm}
\end{table}

\section{Results}
\textbf{Visual Question Answering: } We provide results on unsupervised (see table \ref{tab: res_unsup_vqa}) as well as semi-supervised setups (see table \ref{tab: res_sup_vqa}). We observe that our method outperforms all early exit methods in terms of accuracy and speedup even with less number of trainable parameters. We even outperform the vanilla BLIP-2 inference due to overthinking in the BLIP-2 backbone which is mitigated by our input-adaptive inference. We also provide results on an unsupervised visual dialogue dataset where the task is similar to VQA but there is an additional context before the question i.e. a dialogue history between the user and the model. 

\textbf{Image Captioning:} We provide results of semi-supervised and unsupervised setup in table \ref{tab: res_unsup_cap} and \ref{tab: res_sup_coco} respectively. We clearly outperform the existing models in terms of both accuracy as well as speedup. For the NoCaps dataset, the model is fine-tuned on the COCO dataset. The speedup for NoCaps dataset is lower as there is a domain change from the training which lowers the confidence in prediction taking more samples to deeper exits for inference.

We observe performance improvement over previous baselines as we attached exits to the BLIP-2 model and by performing input-adaptive inference, we perform better than the BLIP-2 model, and as BLIP-2 outperforms other models, \algo{} also outperforms others. For the early exiting baselines on BLIP-2, we outperform them as we have an additional component in the exits rather than just a linear classifier which helps in better performance of exits in terms of both performance and speedup. Note that there is a decrease in accuracy when we are in an unsupervised setup, as our model mimics the final layer hence some amount of overthinking still remains. Still, we are comparable to the BLIP-2 inference. On the other hand, the labeled dataset in semi-supervised tasks helps the model learn the hardness of the incoming sample. This helps the model to overcome the overthinking issue.

We have utilized two versions of the BLIP-2 model that have decoder as FlanT5\textsubscript{XL} and OPT\textsubscript{2.7B}. We observe that the speedup in BLIP-2 with OPT\textsubscript{2.7B} was higher as there are more layers in this hence they are more susceptible to overthinking issues. The speedup for VQA tasks was higher as these tasks are simpler than image captioning tasks. We have not reported the speedup of the models other than the variants of BLIP-2 as they have different types of architectures. In terms of speedup, our objective is to make BLIP-2 faster.

In table \ref{tab: capfilt}, we have shown the result of using the CapFilt method to generate synthetic captions in the absence of the labeled dataset. We have reported the CIDEr score over the NoCaps dataset. We can observe that the model has improved upon the performance using CapFilt and the speedup has significantly increased. This effect is due to the good quality captions that help the exits learn better, hence it outputs more samples early increasing the speedup as compared to knowledge distillation.



\section{Conclusion}
In this study, we introduced a novel inference technique \algo{}, which leverages adversarial training of exits alongside the zero-shot capabilities of the VLMs. By employing \algo{}, we effectively reduce the dependency on a vast amount of labeled training data typically required for exit training. Our approach involves adversarially training exits to generate representations similar to those of the final layer, thereby minimizing the need for extensive labeled data. Moreover, our exit design reduces the number of trainable parameters, resulting in lower computational costs. Experimental results demonstrate that our method significantly enhances inference speed while yielding high-quality outputs.


\section{Limitations}
For attaching exits to a large model such as BLIP-2, the crucial part is to decide where to attach exits within a given budget, i.e., what could be the best places for an exit in the LM component of the backbone without exceeding a certain amount of parameters. We answered that question by explaining the mid-crisis. However, the placements of exits with given budget criteria still remain unexplored, which can make these models even faster within computational boundaries.

\section*{Acknowledgements}
Divya Jyoti Bajpai is supported by the Prime Minister’s Research Fellowship (PMRF), Govt. of India.  Manjesh K. Hanawal thanks funding support from SERB, Govt. of India, through the Core Research Grant (CRG/2022/008807) and MATRICS grant (MTR/2021/000645), and DST-Inria Targeted Programme.  We also thank funding support from Amazon IIT-Bombay AI-ML Initiative (AIAIMLI).

\bibliography{latex/custom}
\newpage
\appendix
\newpage
\section{Appendix}
\label{sec:appendix}
\subsection{Ablation Study}

\textbf{Qualitative analysis:} In Figure~\ref{fig:captions}, we provide some examples of the output provided by the \algo{} model. The figure shows how the early exit models increase the speedup by predicting easier tokens earlier. For instance, the image in the last row and last column of the figure is an example of an easy sample where the tokens are predicted at initial layers. Similarly, for the image with a zebra, it can easily predict the easier token such as `\textit{A zebra standing in a snowy field}' at the initial layers while the part of the image that is not easy to predict `\textit{with a wall behind}' is predicted at deeper layers and predicting a high-quality caption overall while speeding up inference using the easiness of sample as well as token. 

Also note that the common tokens are mostly predicted from the initial layers while the rare tokens are predicted from the deeper layers. This observation suggests that the tokens that occur more number of times are considered as easy by the model while the token that have rare appearances are the ones treated as hard.

\begin{table*}[]
\small
\centering
\begin{tabular}{lccccc}
\hline
Model      & \multicolumn{4}{c}{COCO Karpathy test split} & Speed \\ \hline
           & BLEU4    & CIDEr    & S       & M      &       \\ \hline
MiniGPT    & 38.6     & 133.9    & 24.2    & 30.8   & 1.00$\times$     \\ \hline
DeeMini    & 31.4     & 107.0      & 20.1    & 25.9   & 1.42$\times$  \\
PABEE-Mini & 33.5     & 117.3    & 20.9    & 27.3   & 1.40$\times$   \\
LeeMini    & 37.3     & 126.8    & 22.8    & 29.1   & 1.49$\times$  \\
MuE        & 37.8     & 129.5    & 23.2    & 29.7   & 1.52$\times$  \\ \hline
FREE       & \textbf{38.3}     & \textbf{132.2}    & \textbf{23.9}    & \textbf{30.5}   & \textbf{1.67}$\times$  \\ \hline
\end{tabular}
\caption{Results of MiniGPT-4 model on the COCO Karpathy test split.}
\label{tab:res_minigpt}
\end{table*}

\begin{table}[]
\small
\centering
\begin{tabular}{ccccc}
\hline
\textbf{Noise}  & \textbf{BLEU-4} & \textbf{METEOR} & \textbf{CIDEr} & \textbf{Speed}  \\
\hline
\multicolumn{5}{c}{$\sigma=0.0$}              \\
\hline
BLIP-2 & 40.4   & 29.1   & 141.5 & 1.00$\times$ \\
BLIPEE & 40.5   & 29.1   & 142   & 1.77$\times$ \\
\hline
\multicolumn{5}{c}{$\sigma=0.5$}              \\
\hline
BLIP-2 & 38.6   & 28     & 137.9 & 1.00$\times$ \\
BLIPEE & 39.8   & 28.8   & 139.5 & 1.64$\times$ \\
\hline
\multicolumn{5}{c}{$\sigma=1.0$}              \\
\hline
BLIP-2 & 35.3   & 26.4   & 130.8 & 1.00$\times$ \\
BLIPEE & 36.7   & 27.9   & 134.2 & 1.59$\times$ \\
\hline
\multicolumn{5}{c}{$\sigma=2.0$}                \\
\hline
BLIP-2 & 28.6   & 20.5   & 112.6 & 1.00$\times$ \\
BLIPEE & 31.1   & 22.7   & 120.8 & 1.45$\times$\\
\hline
\end{tabular}
\caption{Results on BLIP-2 and \algo{} on COCO dataset when some level of noise $\sigma$ is added into the images during inference phase.}
\label{tab: robustness}
\end{table}

\begin{table*}[]
\small
\centering
\begin{tabular}{lccccc}
\hline
Model        & \multicolumn{4}{c}{COCO Karpathy test split} & Speed \\ \hline
             & BLEU4    & CIDEr    & S       & M      &       \\ \hline
InstructBLIP & 42.6     & 140.5    & 24.3    & 31.2   & 1.00$\times$     \\ \hline
DeeIB        & 35.2     & 123.7    & 20.8    & 26.8   & 1.39$\times$  \\
PABEE-IB      & 36.5     & 126.3    & 21.7    & 28.0     & 1.35$\times$  \\
LeeIB        & 38.0       & 132.8    & 22.9    & 29.9   & 1.47$\times$  \\
MuE          & 39.4     & 136.2    & 23.4    & 30.6   & 1.42$\times$  \\ \hline
FREE         & \textbf{42.1}     & \textbf{138.9}    & \textbf{23.8}    & \textbf{31.0}     & \textbf{1.58}$\times$  \\ \hline
\end{tabular}
\caption{Results of InstructBLIP model on the COCO Karpathy test split.}
\label{tab:res_instruct}
\end{table*}

\begin{table}[]
\small
\centering
\begin{tabular}{lcccc}
\hline
Model      & VQA  & OKVQA & GQA  & Speed \\ \hline
MiniGPT    & 68.6 & 65.2  & 41.9 & 1.00$\times$     \\ \hline
DeeMini    & 59.8 & 59.1  & 35.7 & 1.53$\times$  \\
PABEE-Mini & 61.4 & 60.8  & 36.9 & 1.49$\times$  \\
LeeMini    & 65.7 & 63.2  & 39.0   & 1.61$\times$  \\
MuE        & 67.8 & 64.6  & 39.9 & 1.59$\times$  \\ \hline
FREE       & \textbf{68.2} & \textbf{65.1}  & \textbf{41.6} & \textbf{1.63}$\times$  \\ \hline
\end{tabular}
\caption{Results on MiniGPT for Visual Question Answering tasks.}
\label{tab:res_minigpt_vqa}
\end{table}

\begin{figure*}
    \centering
    \begin{subfigure}{0.45\textwidth}
        \includegraphics[scale=.35]{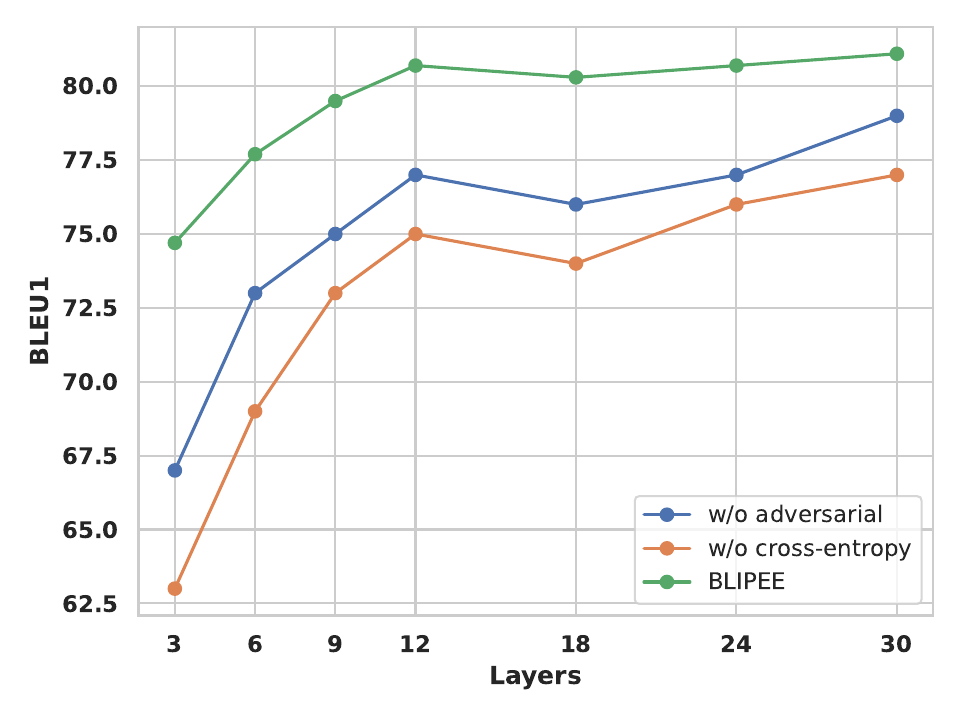}
        \caption{Different combinations of loss function}
        \label{fig:diff_comp}
    \end{subfigure}
    \begin{subfigure}{0.45\textwidth}
        \includegraphics[scale=.35]{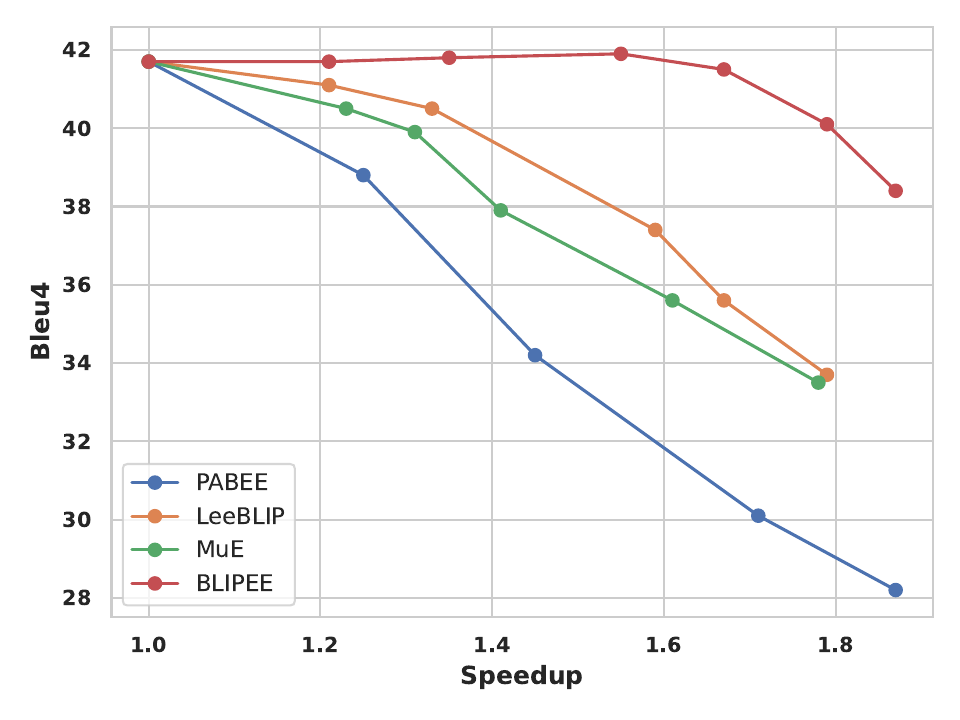}
        \caption{Speedup vs accuracy curve.}
        \label{fig: speedup}
    \end{subfigure}
    \caption{Left: BLEU-1 score for COCO with model BLIP-2-ViT-L-OPT\textsubscript{2.7B} with different components. Right: Speedup vs BLEU-4 curve for COCO dataset with ViT-L-OPT\textsubscript{2.7B}.}
\end{figure*}

\begin{table*}[]
\centering
\small
\begin{tabular}{c|ccccc}
\hline
\multirow{2}{*}{Model} & \multicolumn{5}{c}{No Caps Zero-shot}                      \\
                       & in-domain & near-domain & out-domain & full-dataset & Spd  \\ \hline
w/o CapFilt            & 122.3     & 118.9       & 123.1      & 120.7        & 1.45$\times$\\
Capfilt                & 123.5     & 120.4       & 124.7      & 122.0        & 1.77$\times$ \\ \hline
\end{tabular}
\caption{Difference between CapFilt and Knowledge distillation method for an unsupervised setup.}
\label{tab: capfilt}
\end{table*}

\begin{table*}[]
\centering
\small
\begin{tabular}{c|cc}
\hline
LLM                 & OPT                           & FlanT5                      \\ \hline
Exit Config         & {[}3, 6, 9, 12, 24, 27, 30{]} & {[}3, 5, 7, 9, 12, 20, 22{]} \\
AdamW beta          & {[}0.9, 0.999{]}              & {[}0.9, 0.999{]}            \\
Threshold           & 0.8                           & 0.8                         \\
Inference beam size & 5                             & 5                           \\
Warmup Steps        & 500                           & 500                         \\ \hline
\end{tabular}
\caption{More hyperparameter details of \algo{} on different LM component in the BLIP-2 model. Note that the thresholds are chosen from the set $\{0.5, 0.6, 0.7, 0.8, 0.9, 1.0\}$}
\label{tab: hyperparms detail}
\end{table*}

\begin{table*}[]
    \centering
    \small
    \renewcommand{\arraystretch}{1.2} 
    \setlength{\tabcolsep}{8pt} 
    \begin{tabular}{lccccccc}
        \toprule
        \textbf{Dataset} & \textbf{Layer 3} & \textbf{Layer 6} & \textbf{Layer 9} & \textbf{Layer 12} & \textbf{Layer 24} & \textbf{Layer 27} & \textbf{Layer 30} \\
        \midrule
        VQA-v2  & 0.02 & 0.07 & 0.13 & 0.09 & 0.15 & 0.18 & 0.17 \\
        Ok-VQA  & 0.03 & 0.09 & 0.11 & 0.09 & 0.13 & 0.15 & 0.18 \\
        VizWiz  & 0.01 & 0.06 & 0.09 & 0.08 & 0.11 & 0.17 & 0.22 \\
        \bottomrule
    \end{tabular}

    \caption{Fraction of samples exiting from different layers of the backbone across various datasets.}
    \label{tab:fraction_of_samples}
\end{table*}

\subsection{Accuracy vs speedup}\label{sec: alpha}
In figure \ref{fig: speedup}, we show the accuracy vs speedup curve which could be obtained by changing the threshold parameter $\alpha$. As we decrease the threshold parameter, samples exit from the initial layers even with less confidence, in this way all the samples are more prone to be incorrect decreasing the accuracy but as the threshold is decreased more samples exit from the initial layers and increase the speedup. One key observation is as we start decreasing the threshold, we observe that sometimes the performance even increases, this is the effect of overthinking, where some samples are correctly predicted at initial layers and might become wrong as they reach the final layer. We have also plotted the curves for other exiting methods and observed that our method has better stability as compared to other early exiting methods.
\subsection{Importance of different components}\label{sec: diff_com}
In figure \ref{fig:diff_comp}, we show the importance of different components of our method. We observe that there is a huge performance drop if we remove the knowledge distillation or cross-entropy loss from the overall loss function. This occurs due to catastrophic forgetting or mode collapse where the model gets stuck into local minima. On the other hand, if we remove the adversarial training part, there is again a performance drop, as we only train the classifier but we are not mapping the feature representations of the final layer and the exits hence exits only have low-level features which is insufficient to make correct predictions, hence resulting in a performance drop.

\subsection{Robustness of \algo{}}\label{sec: robustness}
In table \ref{tab: robustness}, we provide results to prove the robustness of \algo{}. To obtain these results, we perform inference on images with Gaussian noise $\sigma$ added to it. The higher the value of $\sigma$, the more noise is present in the image. Observe from the table that when the level of noise is increased in the image the performance of the BLIP-2 model is affected. However, the impact of the noise is smaller for \algo{}.

The reason for the smaller impact of noise on \algo{} is that \algo{} uses outputs from multiple exits and performs inference only when a classifier is confident enough. This makes it robust to noise present during the inference phase.

\section{Results on MiniGPT}\label{sec:MiniGPT}
In Table \ref{tab:res_minigpt} and \ref{tab:res_minigpt_vqa}, we show the results of \algo{} on the MiniGPT backbone. First note that the performance of MiniGPT is not as good as the BLIP-2 model. The reason is stated in the Appendix of the MiniGPT-4 paper as the number of training parameters of the model is significantly lower than the full BLIP-2 model. However, it has a better performance over long sequence generation tasks. We use the Llama \cite{touvron2023llama} model in its decoder.

From the results, we observe that our method can be well generalized over MiniGPT as the results are similar to the BLIP-2 backbone. Note that during fine-tuning MiniGPT, we have used similar hyperparameters as used to train BLIP-2 as the overall architecture of MiniGPT is similar to BLIP-2 except for the number of training parameters. Note that the baselines are similar to those given in the main body with DeeBERT replaced as DeeMini, PABEE as PABEE-Mini, and LeeBERT as LeeMini. All the hyperparameters were kept the same for all the baselines with their approach applied.

\section{Results on InstructBLIP}\label{sec:InstructBLIP}
InstructBLIP is an upgraded version of the BLIP-2 model where the underlying architecture is same but the difference is with the instruction given to generate the output. InstructBLIP can perform better than BLIP-2 over long sequence generation tasks however takes a slight hit on the performance of the image captioning tasks. The hyperparameters for fine-tuning InstructBLIP were the same as for BLIP-2.

However, the behavior of our method applied to this model is the same as it shows minimal performance drop as compared to other EE methods while keeping the speedup better than them. This is the result of the access to deeper level knowledge as our method tries to mimic the behavior of the final layers at the exits. Also, we have used only the validation split to train the exits which makes our method less resource-heavy. 
Note that the baselines are similar to those given in the main body with DeeBERT replaced as DeeIB, PABEE as PABEE-IB, and LeeBERT as LeeIB. All the hyperparameters were kept the same for all the baselines with their approach applied.

Note that both MiniGPT and InstructBLIP share same architectural details as BLIP-2 model with different training strategies, hence we do not explicitly explain the architectural details. Also the training and inference procedure with hyperparameters is similar to the BLIP-2 model.

\section{Analysis on computational cost}\label{sec:analysis_on_computations}
\textbf{Training stage:} In our method, the fine-tuning stage is similar to vanilla fine-tuning of the backbone. After this step, we have additional parameters that are used at the exits consisting of one transformer layer similar to the decoder of the model. While other methods use a classifier at all the exits, note that due to huge size of the vocabulary the classifier size is very large even larger than one transformer layer. Due to this our method has lesser number of training parameters. Now as we train in a GAN-based setup, we alternatively train the exit transformer layer, where it is trained to generate similar representations as the final layer. The generator is the exit transformer layer and a separate feature classifier is also added with a job to discriminate if the features are from the final layer or exit layer. \textbf{The full training procedure along with fine-tuning takes approximately $26$ hours to complete on a setup of $6$ GPUs with $4$ NVIDIA RTX 3090 and $2$ NVIDIA RTX A6000 GPUs while on a similar setup training of vanilla BLIP-2 training requires $18$ hours of training.}

\begin{table*}[]
    \centering
    \small
    \begin{tabular}{l c c}
        \toprule
        \textbf{Component} & \textbf{Parameters per Layer} & \textbf{Total Parameters(32-layers)} \\
        \midrule
        Transformer Layer & $\sim$62.8M & $\sim$2.01B \\
        Layer Norms \& Embeddings & - & $\sim$560M \\
        LM Head & - & $\sim$128.7M \\
        \midrule
        \textbf{Total} & - & \textbf{2.7B} \\
        \bottomrule
    \end{tabular}
    \caption{OPT-2.7B Parameter Breakdown}
    \label{tab:opt2.7b_params}
\end{table*}
\textbf{Inference:} While we have a heavy training setup, post deployment the inference process is very simple and fast. Once the image has been passed through the image encoder as well as the query generator. After this step the query embedding are passed through decoder after the projection layer. Then the decoder generates the text autoregressively. The next token is generated at one of the attached exits where the input to the decoder is processed sequentially at the exits. If an exit is confident on its prediction, it stops the inference process by assigning the token generated by it as the next predicted token by the model. This improves the speed benefits of the full model where not all tokens are processed till the final layer. We observed that more common tokens usually exit at earlier exits while rare tokens are processed deeper into the backbone. As formation of sentence require a large number of common tokens such as `the', `and', etc., the early exits makes the generation process significantly faster. The fraction of token exited from different layers is given in Table \ref{tab:fraction_of_samples}. \textbf{The inference time over the Karpathy test split of the COCO dataset was $7$ minutes of wall clock time approximately while for vanilla BLIP-2, it was approximately $11$ minutes. This also proves the importance of exits.}

In summary, during training, we have additional exits as compared to vanilla BLIP-2 model as well as feature classifiers at the exits. The feature classifiers are dropped during inference and are not further required. While the attached exits help the model to make adaptive predictions by performing inference using some fraction of layers from the full model based on the input sample and token complexity.

\subsection{OPT-2.7B Architecture}\label{sec:parameter_count}

OPT-2.7B is a decoder-only transformer model with 2.7 billion parameters. Each transformer layer consists of multi-head self-attention, a feed-forward network (FFN), and layer normalization. The detailed architecture is as follows:

\begin{itemize}
    \item \textbf{Hidden Size ($d_{\text{model}}$)}: 2560
    \item \textbf{Number of Layers}: 32
    \item \textbf{Attention Heads}: 32
    \item \textbf{FFN Inner Size}: $4 \times d_{\text{model}} = 10240$
    \item \textbf{LM Head Size}: $2560 \times 50272 \approx 128.7$M parameters
\end{itemize}

The total parameter breakdown is summarized in Table~\ref{tab:opt2.7b_params}. Now if we use classical exit setup, that uses independent exit classifiers for attaching exits it adds $128.7M$ parameters at every exit. While in our setup, we attach exit transformers and exit classifiers at intermediate layers where the exit classifier have shared weights as final layer classifier and is frozen. Since the exit transformer is a replica of transformer layer of the model, its adds only $62.8M$ parameters. This reduces the training parameter size by $52$\%.

\end{document}